\documentclass[runningheads]{llncs}
\usepackage{microtype}
\usepackage{graphicx}
\usepackage{subfigure}
\usepackage{booktabs} 
\usepackage{dirtytalk}
\usepackage[T1]{fontenc}
\usepackage{graphicx}
\begin{document}
\title{Can collaborative learning be private, robust and scalable?}
%
%
\author{Dmitrii Usynin \inst{1,2,3} \and Helena Klause \inst{1} \and Johannes C. Paetzold \inst{3,4} \and Daniel Rueckert \inst{1,3} \and Georgios Kaissis \inst{1,2}}
%
\authorrunning{Usynin et al.}
%
\institute{Artificial Intelligence in Medicine and Healthcare, Technical University of Munich, Munich, Germany \and
Institute of Diagnostic and Interventional Radiology, Technical University of Munich, Munich, Germany \and
Department of Computing, Imperial College London, London, United Kingdom \and Helmholtz Zentrum München, Munich, Germany\\
\email{g.kaissis@tum.de}}
\maketitle              
\begin{abstract}
In federated learning for medical image analysis, the safety of the learning protocol is paramount. Such settings can often be compromised by adversaries that target either the private data used by the federation or the integrity of the model itself. This requires the medical imaging community to develop mechanisms to train collaborative models that are private and robust against adversarial data. In response to these challenges, we propose a practical open-source framework to study the effectiveness of combining differential privacy, model compression and adversarial training to improve the robustness of models against adversarial samples under train- and inference-time attacks. Using our framework, we achieve competitive model performance, a significant reduction in model's size and an improved empirical adversarial robustness without a severe performance degradation, critical in medical image analysis.
\end{abstract}

\section{Introduction}
Collaborative machine learning (CML), and in particular collaborative medical image analysis, can significantly benefit from A) having access to large, well-descriptive datasets, which are often highly sensitive and hence difficult to obtain and B) deep machine learning models, which can require significant computational resources during training \cite{rieke2020future,sheller2020federated}. Such models are often trained in a distributed manner, allowing a federation of clients to obtain a joint model without the need to share the data directly, often at the cost of an additional communication burden being put on the federation \cite{reisizadeh2020fedpaq}. The existing methods of collaborative training, such as federated learning, are also particularly vulnerable to inference as well as model poisoning attacks \cite{usynin2021adversarial}, additionally requiring formal means of privacy and integrity protection \cite{kaissis2020secure}. One such scenario was demonstrated by \cite{kaissis2021end}, showing that without carefully selected privacy parameters, the adversary in the context of multi-institutional federated learning on pneumonia classification data was able to reconstruct the private chest X-ray data. Current methods that aim to resolve these issues can pose additional challenges to the federation as they can be difficult to implement in practice (such as model compression, which often requires a public dataset that comes from the same distribution as the training data for calibration), rely on unobvious additional hyper-parameters (such as $\varepsilon$ in DP) or only mitigate a subset of attacks (such as adversarial training that improves model robustness, but does not mitigate any inference attacks). In this work we propose a framework for training and evaluation of ML models, which can help the medical imaging community to A) reduce the communication overhead, B) formally preserve privacy and C) achieve better adversarial robustness. We investigate this by studying model poisoning attacks \cite{fgsm} and their mitigations through the utilisation of differentially private training (DP) \cite{Dwork2013}, model quantization and adversarial training. We investigate two main threat models, which include inference-time and train-time attackers on collaborative learning. Our contributions can be summarised as follows:
\begin{itemize}
    \item We determine how techniques for private and scalable ML (such as DP and model compression) training can be combined to improve adversarial robustness in CML;
    \item We evaluate the most commonly used (e.g. projected gradient descent or PGD) as well as the state-of-the-art (e.g. fast adaptive boundary or FAB) adversarial attacks in these settings and show that the combination of these techniques can provide sufficient protection against utility-oriented adversaries;
    \item We propose an updated view on the relationship between these mechanisms and threat modelling, providing recommendations for achieving improved adversarial robustness using these techniques;
    \item Finally, we propose a framework (namely \textit{PSREval}\footnote{Code available at \url{https://github.com/dimasquest/PSREval}}) for private training and evaluation of image classification models trained in low-trust environments.
\end{itemize}

\section{Related work}
\label{related_work}
Several studies have studied the applications of model compression against adversarial samples in CML \cite{feng2020adversarial,zheng2018pgd,khalid2019qusecnets,lin2019defensive,hong2021qu}, however, there is no prior \textit{unified} perspective on whether quantization techniques improve  adversarial robustness against \textbf{all} utility-based attacks. Authors of \cite{ma2021quantization} discover that when the trained model is subjected to train-time attacks (e.g. backdoor attacks), model compression can significantly reduce robustness. Additionally, the work of \cite{gupta2020improved} highlights, that as there exists a number of quantisation strategies (e.g. discretisation, distillation assisted quantization), a large number of such strategies provide the participants with a semblance of robustness. However, authors of \cite{khalid2019qusecnets} and \cite{rakin2018defend} discover that for a number of inference-time poisoning attacks, model compression reduces the effectiveness of most adversaries. This is due to a smaller set of values that the model can utilise compared to its full-precision counterpart, making the attacker use a significantly higher perturbation budget to affect the decision of the model. 
Originally, \cite{lecuyer2019certified} deployed DP as a method to provably certify ML models against adversarial samples of known perturbation budgets. However, this discussion was limited as the noise was applied directly to the training data or to the output of the first model layer, without considering the arguably most widely used application of DP in deep learning, namely DP-SGD \cite{abadi2016deep}. Various other works \cite{phan2020scalable,bu2021practical} discussed how DP-SGD can be augmented or combined with adversarial training for better model robustness, yet none of them made links to model compression before or considered a train-time attacker, which we address in this work. 
Finally, adversarial training is considered to be one of the most successful empirical defence mechanisms against malicious samples \cite{ganin2016domain,shafahi2019adversarial}, but similarly to model compression, its effects when combined with other robustness enhancement methods have not been studied in sufficient detail.
\begin{figure*}[!ht]
    \centering
\resizebox{.85\linewidth}{!}{\includegraphics[]{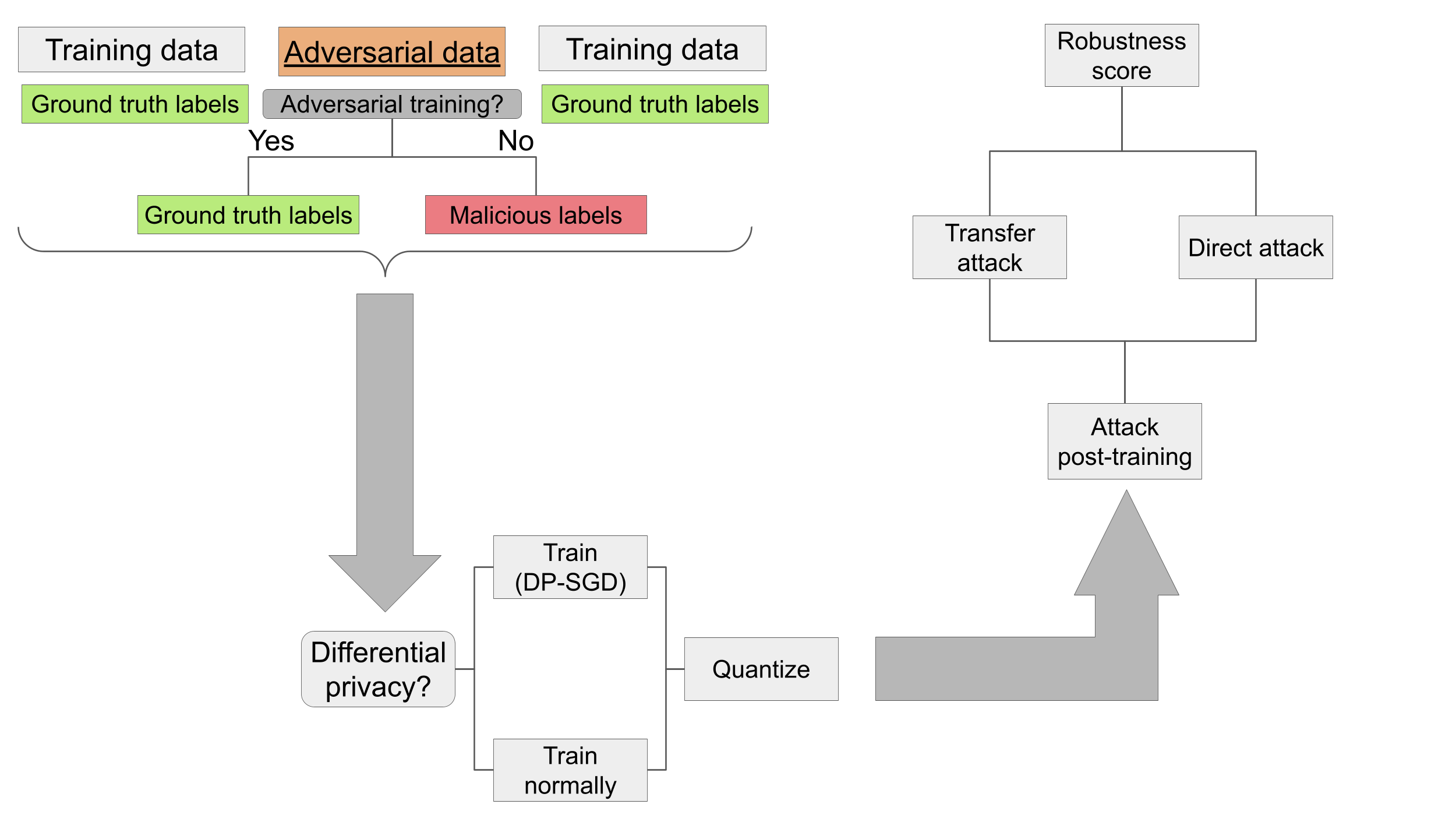}}
    \caption{Overview of our \textit{PSREval} framework, which we describe in Section \ref{methods}.}
    \label{fig:overview}
\end{figure*}

\section{Methods}
\label{methods}
We present an overview of our methodology in Figure \ref{fig:overview}. In this work we generate an adversarial dataset that is used over the course of collaborative training: This dataset can be either used for adversarial training or used to attack the model at train time. We then train two models: One using a normal training procedure and the other using DP-SGD. If the adversary is an active train-time attacker, they would use their malicious data at train time, otherwise the adversary only targets the model at inference-time. Once the models are trained, we then quantize them using the validation dataset (public) to tune the quantization parameters. In this study we perform a \textit{static} model quantization (i.e. post-training quantization of both the weights and the activations), where we replace the full-precision $32$-bit floating point parameters in the model with signed $8$-bit integer parameters. Finally, we perform one of the two attacks to validate the robustness of the model. In the first setting, the adversary has full access (white-box or WB) to the trained model before it is deployed and thus is able to utilise it to generate the adversarial data directly. In the second setting, the adversary is attacking a similar (same architecture, different weights) model that has previously been deployed elsewhere, while only having WB access to the model they obtained during training (partial WB). This setting is termed the \textit{transfer} attack. Our framework provides the robustness scores (in this case accuracies) for both adversarial settings. Note that our framework allows the user to train the models individually and perform model aggregation using their preferred aggregation algorithm (in this study we used federated averaging, where the data was split between $2$ clients, one of which was an adversary).
 
\section{Experiments}
\label{experiments}

\subsection{Experimental setting}
\label{setting}
In this study we perform two collaborative classification tasks on CIFAR-$10$ and paediatric pneumonia prediction (PPPD) (adapted from \cite{kaissis2021end}) datasets. We utilise ResNet-$9$ and ResNet-$18$ architectures. We employ ReLU as our activation function and replace the batch norm layers with group norm layers for compatibility with DP. For DP training we utilise the opacus library \cite{yousefpour2021opacus} with three privacy regimes (representing different end on the privacy-utility spectrum): Concretely, we implemented settings for ($\varepsilon=1.7$), ($\varepsilon=3.4$) and ($\varepsilon=7.0$). For PPPD $\delta=1e^{-4}$ and for CIFAR-$10$ $\delta=1e^{-5}$. We utilise three adversarial attacks methods, namely PGD \cite{pgd}, FGSM \cite{fgsm} and FAB \cite{fab}. When performing train time attacks and adversarial training, we experiment with different proportions of adversarial data, namely $10\%$, $20\%$, $30\%$ or $40\%$ of the training dataset. By default, each attack (if required) is ran for $10$ steps, with a perturbation budget of $8/255$ and a step size (the limit of perturbation during a single step) of $2/255$. We deliberately chose a high perturbation budget (in comparison to the frequently used budget of $2/255$ \cite{fab}) to represent the worst-case scenarios, when the adversary has an ability to significantly affect the training process. We repeat the attacks $10$ times for each setting and report the average values.

\subsection{Performance overview}
We begin by discussing the performance comparison between a normally trained model as well as its DP and quantized counterparts. We present a summary of the standard accuracies for each setting in the Supplementary Material. We note that after the compression procedure, the sizes of ResNet-$9$ and ResNet-$18$ were reduced by $74.4\%$ and $76.6\%$ respectively. The accuracy of the model post-quantization step has not been significantly altered and stayed within $\pm 1\%$ of the original value.
\subsection{Different privacy regimes under quantization}
\begin{figure}[!h]
    \centering
\resizebox{.7\linewidth}{!}{\includegraphics[]{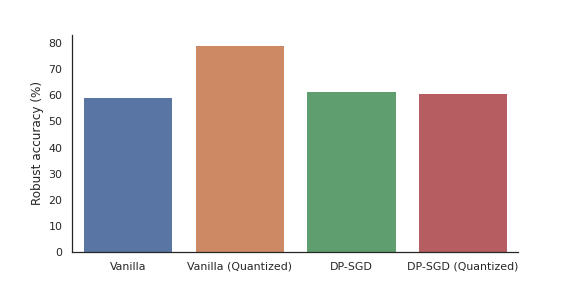}}
    \caption{Transfer attack comparison (generators are the WB models for both, PPPD, ResNet-$18$, $\varepsilon=7.0$). Higher is better. Here we observe that quantization does not affect adversarial robustness of privatised models as much as it affects the non-private models under inference-time attacks.}
    \label{fig:quant_20}
\end{figure}

We experiment with three distinct privacy settings, defined by the values of $\varepsilon$, where lower value represents the \say{stronger} notion of privacy as there is a stricter bound on the release of information content (Section \ref{setting}). This allows us to establish a more clear relationship between the DP-SGD and its ability to affect adversarial robustness when subjected to partial WB attacks. In general, for partial WB attacks, we did not find the DP-trained model to be significantly more robust than the original ones (within $\pm 2\%$), regardless of the privacy regime. When adding post-training quantization, we found that robustness of the model can be improved by up to $5\%$ for smaller models and by up to $20\%$ in larger models (Figure \ref{fig:quant_20}). This seemingly small post-training adaptation allows the federation to achieve a significantly higher adversarial robustness as well as significantly reduce the model size. 

\begin{figure}[!h]
    \centering
\resizebox{.7\linewidth}{!}{\includegraphics[]{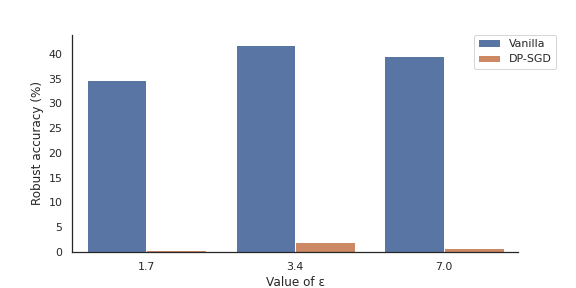}}
    \caption{Transfer attack comparison (generators are the WB models for both, PPPD, ResNet-$18$, $\varepsilon=1.7$). Higher is better. This experiment shows that DP models (of various privacy levels) published online can be used to generate adversarial images for private models of identical architectures with high fidelity.}
    \label{fig:dp_bad_fgsm}
\end{figure}

From Figure \ref{fig:dp_bad_fgsm}, we see that the overall loss of accuracy for DP-trained models (when the adversary uses a model trained with DP-SGD to generate the adversarial samples) is significantly larger than for its non-private counterparts under transfer or partial WB attacks. In fact, we found that the adversary is, in some cases, able to attack a DP-trained model that has the same architecture with almost $100\%$ accuracy, which they are unable to do if the generating model is non-private. The opposite is partially true: If a non-private model is used as a generator, DP-SGD retains a robust accuracy of $30\%$ in comparison to $5\%$ for the original model. Additionally, while for smaller architectures, both the original and the DP models showed a severe lack of robustness, larger models and datasets were significantly more vulnerable only when trained with DP-SGD. This result holds for all three attack implementations and raises questions about the \say{safety} of the publication of private models, because while they provide theoretical guarantees with regards to the privacy of the training data, they can be used as perfect adversarial sample generators, potentially violating the integrity of other learning contexts relying on similar data or architectures. We finally note, that this finding is even more important under the light of the recent publications on the robustness of DP models, as this attack vector has previously not been considered in enough detail, resulting in a semblance of robustness associated with a blind application of DP-SGD without a careful threat model selection.
\subsection{Using adversarial training}
\begin{figure}[!h]
    \centering
\resizebox{.7\linewidth}{!}{\includegraphics[]{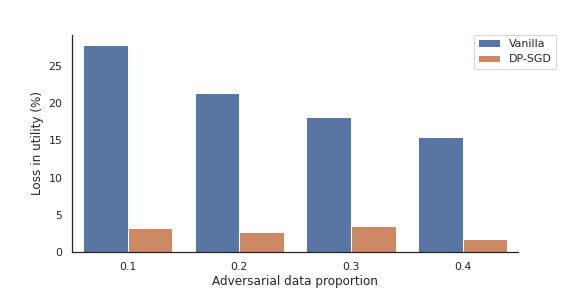}}
    \caption{Accuracy loss under a partially WB attack with adversarial training (generators are the WB models for both, CIFAR-$10$, ResNet-$9$, $\varepsilon=1.7$). Lower is better. Here we see that DP can effectively mitigate a train-time attacker even when they control $40\%$ of the training data.}
    \label{fig:adv_training_high_dp}
\end{figure}

One method that has been particularly effective against utility-based attacks is adversarial training. We analyse three methods of generating the adversarial samples and compare the results to identify the method that is A) effective against malicious adversaries, B) does not result in a significant performance overhead and C) does not interfere with the learning process. We note that FAB was an ineffective method that both severely degraded the performance (a $\times20$ increase in training time) and the utility (down to $20\%$ accuracy in all settings) of the trained model. We see (from Figure \ref{fig:adv_training_high_dp}) that adversarial training can significantly improve the robustness of the trained model in all settings. We also note, however, that this robustness can come at a severe utility cost, which is typically associated with such training process augmentation (reducing the overall accuracy by up to $13\%$ for ResNet-$18$). Similarly to \cite{bu2021practical} we found that adversarial training can be effectively combined with DP training, significantly improving the robustness of the model as well as suffering a much smaller utility penalty when compared to a non-private learning setting. We show exemplary results for a high-privacy (low $\varepsilon$) regime in Figure \ref{fig:adv_training_high_dp} and more in the Supplementary Material. We found PGD to be the optimal sample generation method in a private setting, allowing the federation to mitigate both the privacy-oriented and the utility-oriented attacks. It must be noted, however, that PGD results in a significantly longer training time when compared to FGSM (up to $8$ times longer training for $40\%$ of adversarial samples). In general, we find that there is no \say{optimal} amount of adversarial data that can be used irrespective of the learning context, but using $20\%$ of adversarial data typically resulted in highest robustness across most settings. 
\subsection{Train- and inference-time attacks}
While a number of previous works typically considers an adversary who has a WB access to a pre-trained model, we believe that it is important to evaluate the learning settings against an adversary who actively interferes with the training process itself. As described in our Section \ref{setting}, the adversary controls different proportions of the training data and we study how this can affect the federation. Overall, as seen in Figure \ref{fig:train_time} (as well as in the Supplementary Material), we find that any train time attack can pose a significant risk to a non-private learning setting, irrespective of the dataset or the architecture of the shared model. However, we also found that DP training can \textbf{severely} reduce this risk even for adversaries that control $40\%$ of the training data, as the contributions of the outlier samples are greatly reduced under DP-SGD. This, alongside with the application of adversarial training, leads us to recommend a wider use of DP-SGD against WB attackers. We additionally note that both of these approaches are fully compatible with quantization techniques, allowing the federation to train private and robust models at scale.
\begin{figure}[!h]
    \centering
\resizebox{.7\linewidth}{!}{\includegraphics[]{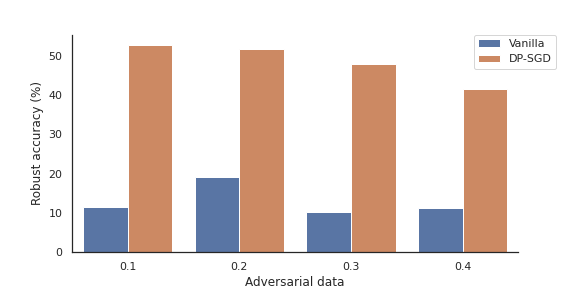}}
    \caption{Robust accuracy under a train-time attacker (CIFAR-$10$, ResNet-$9$, $\varepsilon=3.4$). Higher is better. The result here corresponds to the setting above, showing that we can expect this behavior to hold even under a weaker privacy regime.}
    \label{fig:train_time}
\end{figure}
\section{Discussion and conclusion}
\label{discussion}
In this work we propose a framework for training and evaluation of image analysis models, combining differentially private training, model compression and adversarial training against model poisoning attacks. Our framework allowed us to determine that for the strongest insider adversary, post-training quantization did not have a significant impact on the results of the attack. The opposite is true for partial WB attacks, where the federation enjoys an improvement in robustness of up to $20\%$ in certain contexts compared to an uncompressed setting. In general, we found DP-SGD to be detrimental in partially WB settings, which is primarily due to a significantly lower accuracy of the DP-trained models after training. Our framework revealed that DP-trained models can be more susceptible to transferable adversarial samples. This finding can be surprising, given a \textbf{greatly} higher robustness of DP-trained models (particularly at train time). In essence, DP models are very sensitive to the threat model that the adversary chooses to employ, therefore not allowing a concrete overall conclusion about the effectiveness of this method.
However, there also exist a number of factors that can potentially have an influence on the results of our evaluations that are not explicitly covered. Firstly, similarly to \cite{usynin2022zen}, we discovered that the accuracy of the trained model can have a significant impact on the results of the attack. This is due to the fact that the adversarial labels (i.e. those used by the train-time adversary) are inferred from model predictions and these depend on how well the model is able to distinguish between different classes, affecting the attack. This, in turn, makes it more challenging for us to disentangle how the individual factors that influence model accuracy can affect adversarial robustness. Secondly, in this work, we relied on the post-training quantization, as we find this approach to be the most \textit{practical} (or low-effort and foolproof), as it only requires a single calibration round and a replacement of a small number of operations during model initialisation. Other approaches can be applicable when discussing robustness of collaboratively trained models, such as train-time quantization or quantization-aware training. However, these methods require a larger number of setup steps and adaptations of the training process, making them less practical. 
Finally, there exists a number of contexts that we have not covered in \textit{PSREval}, which go beyond the scope of this work. We are planning to expand our framework with other robustness-enhancement methods, such as adversarial regularisation, knowledge distillation \cite{papernot2017extending} and feature squeezing \cite{xu2017feature}, all of which were previously shown to mitigate utility-oriented adversaries in CML. We used a simple federated averaging aggregated method in our work, therefore leaving more advanced aggregation techniques (some of which can come with additional adversarial robustness) as part of the future work. Additionally, we are aiming to produce a more context-agnostic study, including attacks on image segmentation and object detection tasks, so that the research community can evaluate their model in a much larger number of clinical settings, resulting in a wider adoption of private, robust and scalable training. \\ \\
\textbf{Acknowledgements}
G.K. received funding from the Technical University of Munich, School of Medicine Clinician Scientist Programme (KKF), project reference H14. D.U. received funding from the Technical University of Munich/Imperial College London Joint Academy for Doctoral Studies. This research was supported by the UK Research and Innovation London Medical Imaging and Artificial Intelligence Centre for Value Based Healthcare. J.C.P. was supported by the DCoMEX project (16HPC010), co-financed by the Federal Ministry of Education and Research of Germany and the EuroHPC JU. The funders played no role in the design of the study, the preparation of the manuscript or the decision to publish.
\newpage

\bibliographystyle{splncs04}
\bibliography{main}

\end{document}